\documentclass[10pt,twocolumn,letterpaper]{article}

\usepackage{cvpr}      
\usepackage{makecell}
\usepackage{multirow}
\usepackage{booktabs}
\usepackage{colortbl}
\usepackage{xcolor}
\usepackage{listings}

\definecolor{cvprblue}{rgb}{0.21,0.49,0.74}
\usepackage[pagebackref,breaklinks,colorlinks,allcolors=cvprblue]{hyperref}

\def\paperID{10154} 
\def\confName{CVPR}
\def\confYear{2026}

\title{Reading Between the Lines: Abstaining from VLM-Generated OCR Errors via Latent Representation Probes}
\author{
Jihan Yao\textsuperscript{1}\thanks{Work done during an internship at Google.}, 
Achin Kulshrestha\textsuperscript{2}, 
Nathalie Rauschmayr\textsuperscript{2}, 
Reed Roberts\textsuperscript{2}, \\
Banghua Zhu\textsuperscript{1}, 
Yulia Tsvetkov\textsuperscript{1}, 
Federico Tombari\textsuperscript{2} \\
\textsuperscript{1}University of Washington \quad \textsuperscript{2}Google \\
\texttt{jihany2@cs.washington.edu} \quad \texttt{kulac@google.com} \quad \texttt{rauschmayr@google.com}
}

\begin{document}
\maketitle
\begin{abstract}
As VLMs are deployed in safety-critical applications, their ability to abstain from answering when uncertain becomes crucial for reliability, especially in Scene Text Visual Question Answering (STVQA) tasks. For example, OCR errors like misreading ``50 mph'' as ``60 mph'' could cause severe traffic accidents. This leads us to ask: Can VLMs know when they can't see? Existing abstention methods suggest pessimistic answers: they either rely on miscalibrated output probabilities or require semantic agreement unsuitable for OCR tasks. However, this failure may indicate we are looking in the wrong place: uncertainty signals could be hidden in VLMs' internal representations.

Building on this insight, we propose Latent Representation Probing (LRP): training lightweight probes on hidden states or attention patterns. We explore three probe designs: concatenating representations across all layers, aggregating attention over visual tokens, and ensembling single layer probes by majority vote. Experiments on four benchmarks across image and video modalities show LRP improves abstention accuracy by 7.6\% over best baselines. Our analysis reveals: probes generalize across various uncertainty sources and datasets, and optimal signals emerge from intermediate rather than final layers. This establishes a principled framework for building deployment-ready AI systems by detecting confidence signals from internal states rather than unreliable outputs.

\end{abstract}
\section{Introduction}

\begin{figure*}[!t]
    \centering
    \includegraphics[width=1\linewidth]{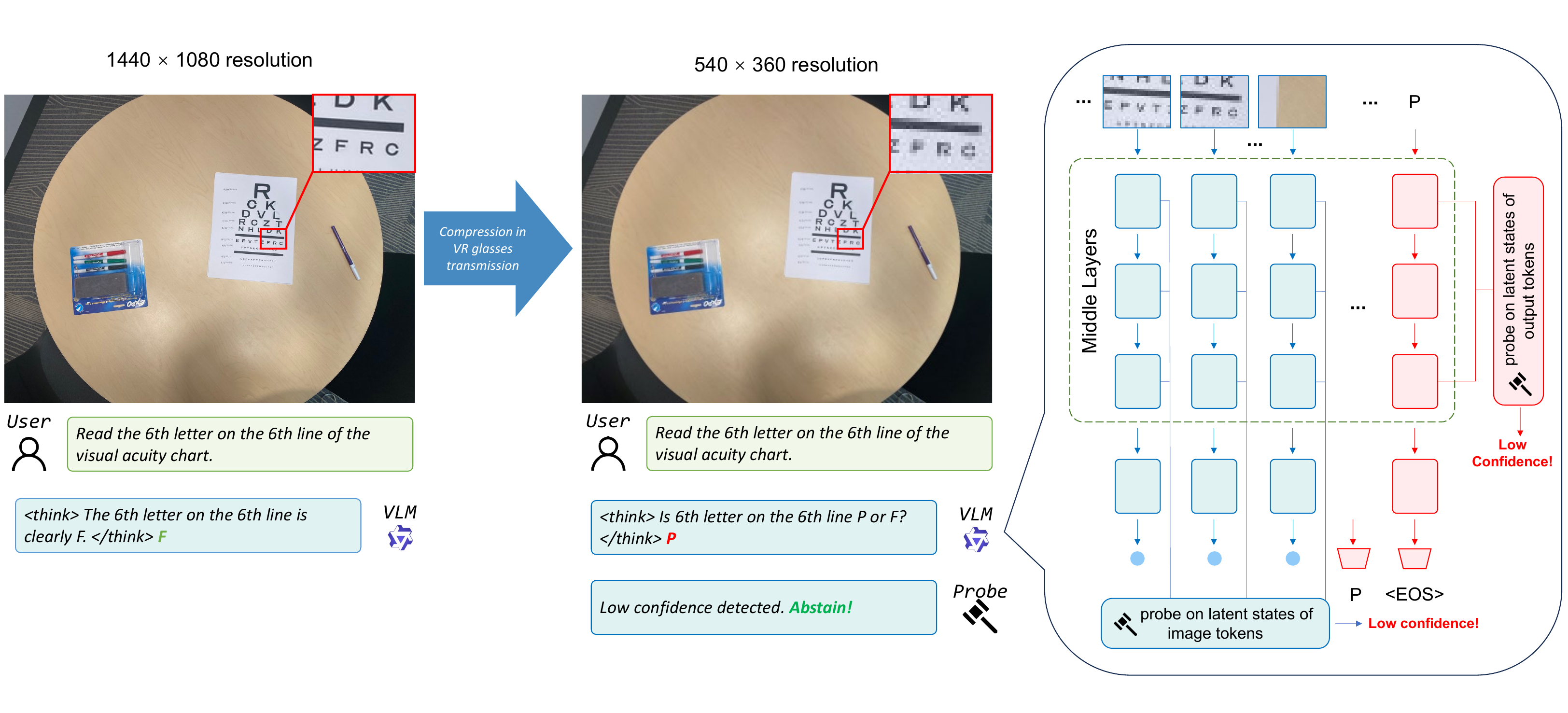}
    \caption{Real-world applications such as assistive AR glasses often encounter low-resolution, corrupted images (middle) rather than the high-quality images (left) typical in laboratory settings. To enable reliable deployment, it is crucial to estimate VLMs' uncertainty and abstain from generating incorrect responses. We propose Latent Representation Probing (LRP) for effective and efficient confidence estimation in VLMs. Our analysis reveals that the most effective probing approaches are: (1) aggregating attention patterns on image token positions across all layers (blue probe, right), and (2) ensembling hidden states from layers that exhibit the strongest confidence signals (red probe, right) through majority vote on output token positions.}
    \label{fig:intro}
\end{figure*}

As Vision-Language Models (VLMs) transition from benchmark testing to operational environments, such as self-driving systems \citep{badue2021self}, robotics \citep{ma2024survey}, and assistive augmented reality glasses \citep{danielsson2020augmented}—ensuring their reliability becomes paramount. In these safety-critical deployments, even minor errors can have severe consequences: misreading "50 mph" as "60 mph" could lead to critical accidents. When models cannot guarantee correctness, \emph{abstention} -- the ability to refrain from answering when uncertain -- serves as a crucial safeguard for preventing such failures.

Among capabilities essential for reliable deployment, Scene Text Visual Question Answering (STVQA) tasks presents particularly reliability challenges. First, OCR is an inherently difficult task for VLMs. VLMs' lightweight alignment layers introduce information bottlenecks that hinder fine-grained perceptual detail transfer \citep{li2023blip}, where image details such as small character recognition are discarded. Studies have found that LLMs using OCR-extracted text significantly outperforming end-to-end VLMs \citep{das2024exams, liu2024ocrbench}. Second, VLMs' OCR capabilities prove especially fragile under real-world conditions: text recognition accuracy deteriorates severely under common corruptions like blur and snow \citep{usama2025analysing}, with performance degrading sharply on imperfect charts and dynamic video environments \citep{wootaek2025losing, nagaonkar2025benchmarking}. Give the difficulties, a central question raises: \emph{Can VLMs know when they can't see?}



Existing uncertainty estimation approaches suggest pessimistic answers. Token-level methods \citep{lin2022teaching, wen2025know} estimate confidence from token probabilities, yet mounting evidence shows that output logits in VLMs are often severely miscalibrated \citep{yao2024varying, feng2024don, xu2025language}. Response-level aggregation methods like self-consistency \citep{wang2022self} and semantic entropy \citep{farquhar2024detecting} are not applicable to free-form answers \citep{chen2023universal} since OCR requires character-level precision instead of semantical equivalence.  However, the failure of these surface-level signals to capture uncertainty does not necessarily mean VLMs lack internal awareness of their perceptual limitations. Instead, it may simply mean we are looking in the wrong place.

We hypothesize that \emph{uncertainty signals exist but are hidden in VLMs' latent representations}. This hypothesis is grounded by two observations: (1) different layers encode qualitatively different information \citep{jawahar2019does, zhang2022dino} with lower layers capturing surface features and upper layers representing more semantic content which may include confidence information, and (2) instruction tuning concentrates changes in final layers \citep{wu2023language} while intermediate layers are better preserved \citep{kossen2024semantic, xiao2025enhancing}. Rather than relying on miscalibrated output logits, we propose \textbf{Probing Latent Representations} -- hidden states and attention patterns -- to uncover the buried uncertainty signals. We explore three probing strategies: (1) concatenating representations across all layers and training a single probe, (2) aggregating attention specifically over visual tokens before probing, and (3) training separate probes for each layer and ensemble by majority vote.

We evaluate our approach across four diverse STVQA benchmarks spanning image and video modalities. Among all methods, ensemble hidden state probing achieves the best performance, improving average abstention accuracy over best baseline by 7.6\%. Analysis reveals that hidden states encode richer semantic uncertainty signals than attention patterns, and LRP is the only method having meaningful abstention performance on video inputs where baselines fail due to their inability to handle long sequences effectively.

To understand the strong performance of hidden state probing, we conduct three analyses. First, we investigate what signals probes capture by evaluating on four synthetic datasets designed to simulate various low model confidence scenarios. The average abstention accuracy of 0.825 suggests they learn general uncertainty representations rather than dataset-specific patterns. Second, layer-wise analysis confirms our hypothesis: the optimal probe operates on intermediate layers instead of the final layer. Finally, to address generalization concerns inherent to training-based methods, cross-dataset evaluation shows that attention pattern probes achieve substantially stronger out-of-distribution performance, improving by 5.8\% over best training-based baselines.

In summary, our contributions are threefold: (1) To the best of our knowledge, this work is the first to evaluate VLMs reliability and abstention capabilities in STVQA tasks and systematically compare various abstention baselines; (2) Our proposed method, latent representation probing, consistently outperforms state-of-the-art baselines across diverse benchmarks; (3) Our method also facilitates comprehensive analyses: we demonstrate that latent representation probing capture robust, generalizable uncertainty representations. This work advances the path toward reliable VLM deployment by enabling models to recognize when they should abstain rather than hallucinate, which is a critical step for trustworthy deployment.

\section{Related Work}

\paragraph{Abstention in VLMs}
Existing approaches of abstention in VLMs can be categorized into three main paradigms. (1) Token-level methods estimate confidence from output probabilities, often applying temperature scaling \citep{tu2024empirical} or verbalized confidence \citep{groot2024overconfidence}. However, these approaches face severe limitations: studies reveal that VLMs exhibit systematic miscalibration with prevalent overconfidence \citep{yao2024varying, feng2024don}. (2) Response-level methods aggregate information across multiple model generations, including self-consistency \citep{wang2022self, zhang2024vl}, semantic entropy \citep{farquhar2024detecting}, and self-evaluation frameworks \citep{chen2024unveiling}. While effective for constrained tasks, these methods struggle with open-ended text recognition where character-level precision is required and semantic similarity is insufficient. (3) Training-based methods fine-tune models to explicitly refuse uncertain queries through refusal-aware instruction tuning \citep{cha2024visually, zhang2024r} or train a better self-reflection model \citep{wu2025generate}. Previous approaches primarily operate on final-layer representations and have not been systematically evaluated on text-intensive VQA tasks.

Parallelly, recent work has explored internal representations for hallucination detection on object-focused VQA tasks. Attention-based approaches \citep{jiang2025devils, zhang2024dhcp, liu2025more, yang2025mitigating} analyze cross-modal attention patterns to detect hallucinations, offering \emph{interpretability} of hallucination. \citet{phukan2024beyond, duan2025truthprint} leverage contextual embeddings from middle layers. These approaches primarily focus on interpretability, aiming to understand \emph{why} hallucinations occur rather than maximizing detection performance. Moreover, some methods have high computational cost, making it prohibitive for video inputs. In contrast, our lightweight probes require only a single forward pass.

\paragraph{STVQA Benchmarks}
STVQA encompasses diverse tasks across both image and video. Image-based benchmarks range from basic scene text understanding (TextVQA \citep{singh2019towards}, ST-VQA \citep{biten2019scene}) to specialized domains including document analysis (DocVQA \citep{mathew2021docvqa}), chart interpretation (ChartQA \citep{masry2022chartqa}), and book covers (OCRVQA \citep{mishra2019ocr}). Video benchmarks extend these challenges with temporal reasoning requirements (M4-ViteVQA \citep{zhao2022towards}, EgoTextVQA \citep{zhou2025egotextvqa}). Comprehensive evaluations like OCRBench \citep{liu2024ocrbench} and SeedBench Plus 2 \cite{li2024seed} assess models across key information extraction, text location, etc. These benchmarks collectively evaluate models' abilities in spatial and temporal reasoning, and fine-grained perception. However, STVQA presents unique challenges: text appears in arbitrary orientations and fonts, and usually come up with corruptions, demanding robust performance under diverse real-world conditions.

\section{Baselines}
Despite growing interest in VLM reliability, existing abstention methods for STVQA tasks lack systematic comparison. To establish a comprehensive evaluation of these approaches, we define the abstention problem as: given a visual input $\mathbf{v}$ (image or video) and a question $\mathbf{q}$, a generated answer $\mathbf{a} = \text{VLM}(\mathbf{v}, \mathbf{q})$, our goal is to determine when the model should abstain from answering. We aim to learn a confidence score $s(\mathbf{v}, \mathbf{q}, \mathbf{a}, \mathbf{e}) \rightarrow [0, 1]$ that estimates the model's uncertainty, where $\mathbf{e}$ is optional external confidence evidence. The model should abstain when $s(\mathbf{v}, \mathbf{q}, \mathbf{a}, \mathbf{e}) < \tau$ and answer otherwise, where $\tau$ is a selected threshold on the validation set.


\paragraph{Token Probability} A well-calibrated VLM's token probabilities should reflect response reliability \citep{geng2023survey}, with more probable answers more likely correct. We use the average token probability as the confidence indicator. Specifically, for an answer $\mathbf{a} = [t_1, t_2, \ldots, t_n]$ with token probabilities $[p_1, p_2, \ldots, p_n]$, the abstention function is:
\begin{equation}
    s = \exp(\frac{1}{n}\sum_{i=1}^{n} log(p_i))
\end{equation}

\paragraph{Ask for Calibration} Verbalized confidence has been proven to be more calibrated than token probability \citep{tian2023just}. After generating answer $\mathbf{a}$, we elicit a numerical confidence $c \in [0, 1]$ by prompting \textit{``Rate how likely the answer is correct from 1 to 5, where 5 is the most likely.''}:

\begin{equation}
    s = \text{VLM}([\mathbf{v}, \mathbf{q}, \mathbf{a}, \text{PROMPT}])
\end{equation}
, where $[\cdot]$ denotes concatenation.

\paragraph{Self Consistency} Self-consistency~\citep{wang2022self} generates multiple responses and uses majority voting for abstention. For OCR tasks with free-form responses, we measure answer similarity using character accuracy (CA). Given $k$ sampled answers $\{\mathbf{a}_1, \ldots, \mathbf{a}_k\}$, the confidence is estimated by:
\begin{equation}
    s = \frac{1}{k}\sum_{i=1}^{k} \text{CA}(\mathbf{a}, \mathbf{a}_i)
\end{equation}

\paragraph{Prompt to Abstain} Inspired by none-of-the-above options in multiple-choice QA~\citep{kadavath2022language}, we adapt this approach for free-form OCR tasks by instructing the model to abstain when uncertainty. We prepend the prompt with: \textit{``Answer `I don't know' if you are uncertain about your answer.''} The confidence function is then:
\begin{equation}
    s = \mathbb{I}(\text{``I don't know''} \notin \mathbf{a})
\end{equation},
where $\mathbb{I}[\cdot]$ is the indicator function.

\paragraph{VLM Judge} We can train a VLM to judge the answer correctness \citep{kadavath2022language, yao2025mmmg}. Given the visual input $\mathbf{v}$, question $\mathbf{q}$, and answer $\mathbf{a}$, along with the instruction \textit{``Is this answer correct? Answer True or False.''}, the VLM judge should output a special token whose probability indicates correctness:
\begin{equation}
    s = \mathbb{I}(p(\text{``True''}|\mathbf{v}, \mathbf{q}, \mathbf{a}) > p(\text{``False''}|\mathbf{v}, \mathbf{q}, \mathbf{a}))
\end{equation}

\paragraph{R-Tuning} Following R-tuning approaches \citep{zhang2024r}, we fine-tune the model to express confidence alongside its answers. During training, we augment correct answers with the suffix \textit{``I am sure''} and incorrect answers with \textit{``I am unsure''}. At inference, the model generates both answer and confidence, and we extract the confidence score:
\begin{equation}
    s = \mathbb{I}(\text{``I am sure''} \in \mathbf{a})
\end{equation}

\paragraph{Summed Visual Attention Ratio (SVAR)} Inspired by the fact that hallucinated answers receive lower overall attention to visual tokens, SVAR \citep{jiang2025devils} uses the image token attention weights from middle layers as confidence signals. The confidence is estimated by:
\begin{equation}
    s = \sum_{\ell=\ell_{\text{start}}}^{\ell_{\text{end}}} \frac{1}{H} \sum_{h=1}^{H} \sum_{i=1}^{n} A_{i, o_1}^{(\ell,h)}
\end{equation},
where $A_i^{(\ell,h)}$ is the attention weight of the first output token's $o_1$ attention on $i$-th visual token at layer $\ell$ and head $h$, $n$ is the number of visual tokens, $H$ is the number of attention heads.

\paragraph{Contextual Lens} Contextual Lens \citep{phukan2024beyond} leverages hidden states from middle layers instead of attention weights. Given image token hidden states $\{\mathbf{h}_i^{(\ell)}\}_{i=1}^{n}$ and answer token hidden states $\{\mathbf{h}_j^{(\ell)}\}_{j=1}^{m}$ at layer $\ell$, the confidence is computed as the maximum cosine similarity between them:
\begin{equation}
    s = \max_{\ell} \max_{i} \frac{\mathbf{h}_i^{(\ell)} \cdot \left(\sum_{j=1}^{m}\mathbf{h}_j^{(\ell)}\right)}{\|\mathbf{h}_i^{(\ell)}\| \|\sum_{j=1}^{m}\mathbf{h}_j^{(\ell)}\|}
\end{equation}
where $m$ is the number of answer tokens at layer $\ell$.

\section{Latent Representation Probing (LRP)}

Our approach trains lightweight probes on VLMs' internal representations to detect OCR errors. We investigate two types of latent representations during inference and three different probe designs.

\subsection{Latent Representations}

\paragraph{Hidden State Representations}
We extract hidden state representations from intermediate transformer layers. For a VLM with $L$ transformer blocks, let $\mathbf{h}_j^{(\ell)}$ denote the latent state of the $j$-th token at layer $\ell$, where $\mathbf{h}^{(\ell)} = \text{TransformerBlock}_\ell(\mathbf{h}^{(\ell-1)})$. Given an answer $\mathbf{a} = [t_1, t_2, \ldots, t_m]$, we average the hidden states of all output tokens at each layer: 
\begin{equation}
\mathbf{h}_{\text{out}}^{(\ell)} = \frac{1}{m}\sum_{j=1}^{m} \mathbf{h}_{t_j}^{(\ell)} \in \mathbb{R}^D
\end{equation}
, where $D$ is the hidden dimension.

\paragraph{Attention Pattern Representations}
For each attention head $h$ at layer $\ell$, the attention weight from output token $j$ to input token $i$ is computed as $A_{j,i}^{(\ell,h)}$. We average the attention weights across all output tokens: 
\begin{equation}
    \mathbf{a}_{\text{out}}^{(\ell)} = \frac{1}{m}\sum_{j=1}^{m} A_{j,:}^{(\ell,:)} \in \mathbb{R}^{n \times H}
\end{equation}, where $m$ is the number of output tokens, $n$ is the number of input tokens, and $H$ is the number of attention heads.

\subsection{Probe Design}

\paragraph{Concatenated Probe}
For latent states, we concatenate and flatten representations across all $L$ layers, then train a 4-layer MLP $f_\theta: \mathbb{R}^{L \cdot D} \rightarrow \{0,1\}$:
\begin{equation}
    s = f_\theta\left([\mathbf{h}_{\text{out}}^{(1)}, \mathbf{h}_{\text{out}}^{(2)}, \ldots, \mathbf{h}_{\text{out}}^{(L)}]_{\text{flatten}}\right)
\end{equation}
For attention patterns, we flatten across layers and heads for each input token position, then train a 4-layer transformer encoder $g_\theta: \mathbb{R}^{n \times (L \cdot H)} \rightarrow [0,1]$:
\begin{equation}
    s = g_\theta\left([\mathbf{a}_{\text{out}}^{(1)}, \mathbf{a}_{\text{out}}^{(2)}, \ldots, \mathbf{a}_{\text{out}}^{(L)}]_{\text{flatten}}\right)
\end{equation}

\paragraph{Visual-Focused Probe}
Motivated by prior work \citep{jiang2025devils} showing that attention to visual tokens is most informative for hallucination detection, we propose using multi-head activation as confidence features. Let $\mathcal{V}$ denote the set of visual token positions. We average attention weights over visual tokens: $\mathbf{a}_{\text{vis}}^{(\ell)} = \frac{1}{|\mathcal{V}|}\sum_{i \in \mathcal{V}} \mathbf{a}_{\text{out}, i}^{(\ell)} \in \mathbb{R}^H$. We then apply a 4-layer MLP $f_\theta: \mathbb{R}^{L \cdot H} \rightarrow \{0,1\}$:
\begin{equation}
    s = f_\theta\left([\mathbf{a}_{\text{vis}}^{(1)}, \mathbf{a}_{\text{vis}}^{(2)}, \ldots, \mathbf{a}_{\text{vis}}^{(L)}]_{\text{flatten}}\right)
\end{equation}

\paragraph{Ensemble Probe}
Instead of using all layers jointly, we train individual probes for each layer separately. For latent states, we train $L$ independent MLPs $\{f_\theta^{(\ell)}\}_{\ell=1}^{L}$, where each $f_\theta^{(\ell)}: \mathbb{R}^d \rightarrow \{0,1\}$ operates on a single layer's representation. We evaluate each layer's performance on a validation set and select the top-$K$ layers with highest abstention accuracy. At inference, we aggregate predictions through majority voting:
\begin{equation}
    s = \mathbb{I}[\frac{1}{K}\sum_{\ell \in \mathcal{L}_{\text{top-K}}} f_\theta^{(\ell)}(\mathbf{h}_{\text{out}}^{(\ell)}) \geq 0.5]
\end{equation}
where $\mathcal{L}_{\text{top-K}}$ denotes the selected layers. In our experiments, we set $K=5$. The same approach applies to attention pattern probes.

\subsection{Training}
The probes are trained using binary cross-entropy loss with soft labels:
\begin{equation}
    \mathcal{L} = -\frac{1}{N}\sum_{i=1}^{N} [y_i \log(s_i) + (1-y_i)\log(1-s_i)]
\end{equation}
where $y_i \in [0, 1]$ is the soft correctness label computed by each dataset's evaluation script and $N$ denotes the total number of samples. Since the probes we train are classifiers, LRP are threshold-free during inference.

\section{Experiments}
\subsection{Models and Datasets}
We employ two open-source VLMs for our experiments, \textsc{Qwen-VL-2.5} \citep{bai2025qwen2} and \textsc{Gemma3-12B} \citep{team2025gemma} respectively. For single-pass methods, we employ greedy decoding and max generation length 128 to ensure reproducible results. We use temperature 1.0 and generate 5 responses per prompt when sampling is required. Since \textsc{Gemma3-12B} doesn't natively support video inputs, we process videos into 1 fps sequences and inputs as multiple images. For more detailed experiment setup, please refer to Appendix.

We conduct experiments on four text-rich VQA datasets spanning both image and video modalities, with the first two focused on general OCR tasks and last two specifically on in-the-wild Text VQA tasks. For all datasets, we split the data into training, validation, and test sets with a 6:2:2 ratio. We select the abstention threshold $\tau$ on the validation set and report all results on the test set.
\begin{itemize}[leftmargin=*]
    \item \textbf{OCRBench v2} \citep{fu2024ocrbench} is by far the most comprehensive OCR evaluation benchmark comprising 10K human-verified QA pairs across text recognition, text location, scene text-centric VQA, document-oriented VQA, text element parsing, and knowledge-intensive text reasoning.
    
    \item \textbf{SEED-Bench-2-Plus} \citep{li2024seed} contains 2.3K multiple-choice questions with human annotations spanning three online visual displays (charts, maps, and webs). We evaluate models using free-form generation instead of the original multiple choices to increase task difficulty.
    
    \item \textbf{HierText} \citep{long2022towards} features hierarchical annotations of text in the wild scenes and documents, containing 11K images with word, line, and paragraph level annotations. Since HierText provides only OCR annotations, we use \textsc{Gemini-2.5-Pro} \citep{comanici2025gemini} to synthesize question-answer pairs based on the annotations, followed by human validation to ensure quality and relevance.
    
    \item \textbf{EgoTextVQA} \citep{zhou2025egotextvqa} is a egocentric video QA benchmark containing 1.5K outdoor driving and 4.4K indoor robotics housekeeping activities. Due to computational limitation, we sample the videos clips with less 1 minute duration.
\end{itemize}

\subsection{Evaluation Metrics}


We adopt four complementary metrics to evaluate abstention performance following previous work \citep{feng2024don, wen2025know}. Since OCR labels are continuous rather than binary, we modify the standard metrics to handle soft correctness labels. 

\paragraph{Effective Reliability (ER)} evaluates the overall system reliability of VLMs with abstention mechanism. It rewards correct answers while penalizing incorrect ones, thus penalizing both over-abstaining and under-abstaining:
\begin{equation}
    \text{ER} = \frac{1}{N}\sum_{i=1}^{N} (2y_i - 1) \cdot \mathbb{I}[s_i \geq \tau]
\end{equation}

\paragraph{Abstention Accuracy (A-Acc)} evaluates whether the model makes correct abstention decisions: abstaining on incorrect answers and answering on correct answers:
\begin{equation}
    \text{A-Acc} = \frac{1}{N}\sum_{i=1}^{N} \left[\mathbb{I}[s_i < \tau] \cdot (1-y_i) + \mathbb{I}[s_i \geq \tau] \cdot y_i\right]
\end{equation}

\paragraph{Reliable Accuracy (R-Acc)} penalizes under-abstention where the model should abstain when they are likely to make critial mistakes:
\begin{equation}
    \text{R-Acc} = \frac{\sum_{i=1}^{N} y_i \cdot \mathbb{I}[s_i \geq \tau]}{\sum_{i=1}^{N} \mathbb{I}[s_i \geq \tau]}
\end{equation}

\paragraph{Abstention Precision (A-Pre)} penalizes over-abstention where the model unnecessarily abstains on samples it could answer correctly:
\begin{equation}
    \text{A-Prec} = \frac{\sum_{i=1}^{N} (1-y_i) \cdot \mathbb{I}[s_i < \tau]}{\sum_{i=1}^{N} \mathbb{I}[s_i < \tau]}
\end{equation}

\subsection{Results}
\begin{table*}[!t]
\centering
\tabcolsep=0.04cm
\small

\begin{tabular}{lcccc|cccc|cccc|cccc|c}\toprule[1.5pt]
\multirow{1}{*}{\textbf{Method}}& \multicolumn{4}{c}{\textsc{OCRBench v2}} &\multicolumn{4}{c}{\textsc{SeedBench Plus 2}} &\multicolumn{4}{c}{\textsc{HierText}} &\multicolumn{4}{c}{\textsc{EgoTextVQA}}&\multirow{3}{*}{Avg}\\
&ER &A-Acc &R-Acc &A-Pre &ER &A-Acc &R-Acc &A-Pre &ER &A-Acc &R-Acc &A-Pre &ER &A-Acc &R-Acc &A-Pre & \\

\midrule[0.75pt]\multicolumn{17}{c}{\textit{\ \ \textbf{\textsc{Qwen2.5-VL-7B}}} }\\ \midrule[0.75pt]
Token Prob. &0.108 &0.665 &0.732 &0.645 &0.169 &0.667 &0.651 &0.687 &\underline{0.161} &0.663 &0.644 &0.685 &0.049 &0.681 &0.622 &0.695 & 0.669\\
Ask for Calib. &0.035 &0.592 &0.560 &0.606 &0.156 &0.654 &\underline{0.784} &0.604 &0.034 &0.536 &0.521 &0.607 &0.001 &0.635 &0.501 &0.731 & 0.604\\
Self Consist. &0.120 &0.677 &\textbf{0.868} &0.640 &0.138 &0.636 &0.658 &0.619 &0.124 &0.625 &\textbf{0.722} &0.642 &0.010 &0.644 &\textbf{0.727} &0.642 & 0.646\\
Prompt to Abs. &-0.164 &0.418 &0.418 &\textbf{1.000} &0.057 &0.529 &0.529 &\textbf{1.000} &-0.146 &0.427 &0.427 &\textbf{1.000} &-0.328 &0.336 &0.336 &\textbf{1.000} &0.427 \\
VLM Judge &0.211 &0.768 &0.755 &0.777 &\underline{0.316} &\underline{0.737} &0.742 &0.728 &\textbf{0.192} &\textbf{0.749} &\underline{0.689} &0.811 &0.000 &0.634 &1.000 &0.634 & 0.722\\
R-Tuning &0.122 &0.680 &0.596 &0.828 &\underline{0.316} &\underline{0.737} &0.709 &0.821 &-0.090 &0.467 &0.453 &0.833 &- &- &- &- & 0.628\\
SVAR &0.030 &0.587 &0.550 &0.604 &0.184 &0.605 &0.644 &0.537 &-0.093 &0.464 &0.448 &0.595 &0.003 &0.636 &0.562 &0.638 & 0.573\\
Context Lens &0.016 &0.573 &0.516 &0.627 &0.162 &0.583 &0.583 &0.625 &-0.003 &0.554 &0.498 &0.684 &0.000 &0.634 &1.000 &0.634 & 0.586\\
\rowcolor{blue!10}Concat (Attn.) &0.036 &0.594 &0.522 &\underline{0.901} &0.300 &0.721 &0.689 &\underline{0.849} &0.071 &0.628 &0.553 &0.783 &0.069 &0.703 &0.599 &0.757 &0.662 \\
\rowcolor{blue!10}Concat (Hid.) &\underline{0.224} &\underline{0.781} &0.774 &0.786 &0.314 &0.735 &0.712 &0.798 &0.121 &0.678 &0.611 &0.757 &0.085 &0.718 &0.611 &\underline{0.785} & \underline{0.728}\\
\rowcolor{blue!10}Visual (Attn.) &0.210 &0.768 &0.760 &0.772 &0.314 &0.735 &\textbf{0.785} &0.673 &0.102 &0.659 &0.575 &\underline{0.837} &\underline{0.088} &\underline{0.722} &0.653 &0.750 & 0.721\\
\rowcolor{blue!10}Ensemble (Attn.) &0.171 &0.729 &0.678 &0.775 &0.246 &0.667 &0.650 &0.744 &0.093 &0.650 &0.660 &0.646 &0.043 &0.676 &\underline{0.713} &0.672 & 0.680\\
\rowcolor{blue!10}Ensemble (Hid.) &\textbf{0.231} &\textbf{0.789} &\underline{0.781} &0.794 &\textbf{0.342} &\textbf{0.763} &0.775 &0.744 &0.152 &\underline{0.709} &0.652 &0.765 &\textbf{0.104} &\textbf{0.738} &0.674 &0.765 & \textbf{0.750}\\

\midrule[0.75pt]\multicolumn{17}{c}{\textit{\ \ \textbf{\textsc{Gemma3-12B}}} }\\ \midrule[0.75pt]

Token Prob. &0.094 &0.700 &0.605 &0.777 &0.136 &0.656 &0.674 &0.644 &0.059 &0.690 &0.583 &0.750 &0.098 &0.595 &\underline{0.597} &0.594 & 0.660\\
Ask for Calib. &-0.169 &0.436 &0.409 &0.806 &-0.024 &0.496 &0.488 &\underline{0.889} &-0.263 &0.368 &0.368 &\textbf{1.000} &0.023 &0.518 &0.512 &\textbf{0.867} & 0.455\\
Self Consist. &0.137 &0.743 &0.718 &0.754 &0.215 &0.735 &0.693 &0.787 &0.139 &\textbf{0.771} &0.680 &0.828 &\underline{0.113} &\underline{0.608} &0.596 &0.626 & 0.714\\
Prompt to Abs. &-0.148 &0.453 &0.421 &\underline{0.951} &0.132 &0.568 &0.566 &\textbf{1.000} &-0.111 &0.449 &0.444 &\textbf{1.000} &0.044 &0.546 &0.523 &\underline{0.866} &0.504 \\
VLM Judge &0.166 &0.760 &\textbf{0.840} &0.735 &0.230 &0.667 &\textbf{0.862} &0.576 &0.068 &0.632 &0.547 &0.846 &0.112 &0.607 &0.576 &0.698 & 0.667\\
R-Tuning &-0.117 &0.477 &0.437 &\textbf{0.955} &0.268 &0.704 &0.675 &0.796 &-0.127 &0.437 &0.437 &\textbf{1.000} &- &- &- &- & 0.539\\
SVAR &0.005 &0.599 &0.514 &0.619 &0.129 &0.566 &0.570 &0.514 &0.003 &0.567 &\textbf{1.000} &0.565 &0.009 &0.505 &0.505 &\textbf{1.000} & 0.559\\
Context Lens &-0.123 &0.471 &0.429 &0.734 &0.101 &0.537 &0.557 &0.389 &-0.025 &0.539 &0.423 &0.561 &0.017 &0.512 &0.508 &0.733 & 0.515\\
\rowcolor{blue!10}Concat (Attn.) &-0.016 &0.578 &0.490 &0.928 &\underline{0.300} &\underline{0.737} &\underline{0.784} &0.684 &0.146 &0.709 &0.658 &0.753 &- &- &- &- & 0.674\\
\rowcolor{blue!10}Concat (Hid.) &0.153 &0.747 &0.669 &0.810 &0.281 &0.717 &0.705 &0.743 &0.046 &0.610 &0.529 &\underline{0.924} &0.108 &0.604 &0.573 &0.689 & 0.669\\
\rowcolor{blue!10}Visual (Attn.) &\underline{0.187} &\underline{0.781} &0.769 &0.788 &0.272 &0.708 &0.731 &0.676 &\textbf{0.158} &\underline{0.721} &0.669 &0.767 &- &- &- &- &\textbf{0.737}\\
\rowcolor{blue!10}Ensemble (Attn.) &0.140 &0.734 &0.724 &0.739 &0.268 &0.704 &0.705 &0.703 &0.149 &0.712 &\underline{0.718} &0.709 &- &- &- &- & 0.717\\
\rowcolor{blue!10}Ensemble (Hid.) &\textbf{0.203} &\textbf{0.797} &\underline{0.792} &0.800 &\textbf{0.311} &\textbf{0.748} &0.728 &0.792 &\underline{0.152} &0.715 &0.637 &0.813 &\textbf{0.152} &\textbf{0.647} &\textbf{0.669} &0.629 & \underline{0.727}\\

\bottomrule[1.5pt]
\end{tabular}
\caption{We evaluate the abstention performance of eight methods and our proposed method on four STVQA datasets across image and video modalities. Best results are in \textbf{bold}, second best are in \underline{underline}, incompatible or unavailable results are denoted as ``-'' and our methods are in \colorbox{blue!10}{blue background}.. ``Avg'' is the micro-average of abstention accuracy of four datasets. Too high reliable accuracy or abstention precision usually indicates over-abstention or under-abstention, while effective reliability and abstention accuracy are comprehensive metrics considering both over abstention or under-abstention. Ensemble probe on hidden states yield the best average abstention accuracy while visual-focused probe on attention pattern yield the second best.}

\label{tab:main}
\end{table*}

We present the abstention performance of baseline methods and LRP in Table \ref{tab:main}.
\begin{itemize}[leftmargin=*]
    \item \textbf{Ensemble hidden state probe and visual-focused attention pattern probe are the best.} For hidden states, the ensemble probe yields the best average effective reliability and abstention accuracy, improving upon the best baseline by 33.8\% and 7.6\% respectively. For attention patterns, the visual-focused probe achieves the second-best performance, improving by 23.3\% and 6.1\%. This suggests attention patterns, which are scalar relevance scores, lack the rich semantic information encoded in hidden states that is necessary for detecting OCR errors.
    
    \item \textbf{Baseline methods fail for various reasons.} Among training-free methods, self-consistency achieves the best performance with 68.0\% average abstention accuracy. However, its high reliable accuracy paired with low abstention precision on \textsc{Qwen2.5-VL-7B} reveals a tendency to over-abstain due to threshold sensitivity. Conversely, prompt-to-abstain exhibits severe under-abstention with extremely high abstention precision, indicating that small VLMs lack self-reflection capabilities. Among training-based methods, LRP surpasses VLM Judge in 7 of 8 (model, dataset) combinations, demonstrating that training lightweight probes on latent representations is more effective and efficient than end-to-end fine-tuning. R-tuning's failure in 2 of 6 combinations further reveals its sensitivity to dataset composition (ratio of certain vs. uncertain examples) and training instability.
    
    \item \textbf{LRP is the only abstention method working for video modality.} On video datasets, LRP is the only method providing meaningful abstention performance, while all baseline methods achieve an average effective reliability of merely 0.051. This advantage arises because video inputs span hundreds of frames, but latent representations compress this information into compact embeddings, making LRP naturally more robust to long sequences compared to baselines.
\end{itemize}

\section{Analysis}
\subsection{Various Abstention}
To verify that our probes learn general uncertainty signals, we create four unseen synthetic dataset from HierText, and measure the zero-shot prediction accuracy of probes trained on \textsc{HierText}. Each dataset simulates a distinct uncertainty source described in \cite{wen2025know}: (1) \textbf{Visual occlusion}: masking text regions with white boxes; (2) \textbf{Visual blur}: adding Gaussian noise to the image ($\sigma=5$); (3) \textbf{Semantic unanwerability}: using \textsc{Gemini-2.5-Pro} to replace a few words in the questions such that they ask about non-existent text or regions; (4) \textbf{Model uncertainty}: prepending a random low-confidence prefix (e.g., "I'm not sure") to model outputs before generation. The intuition behind is: if probes decode genuine uncertainty signals from latent representations, they should generalize across different uncertainty sources, as these perturbations all fundamentally affect model confidence despite having distinct forms. 

\begin{figure}[!t]
    \centering
    \includegraphics[width=1\linewidth]{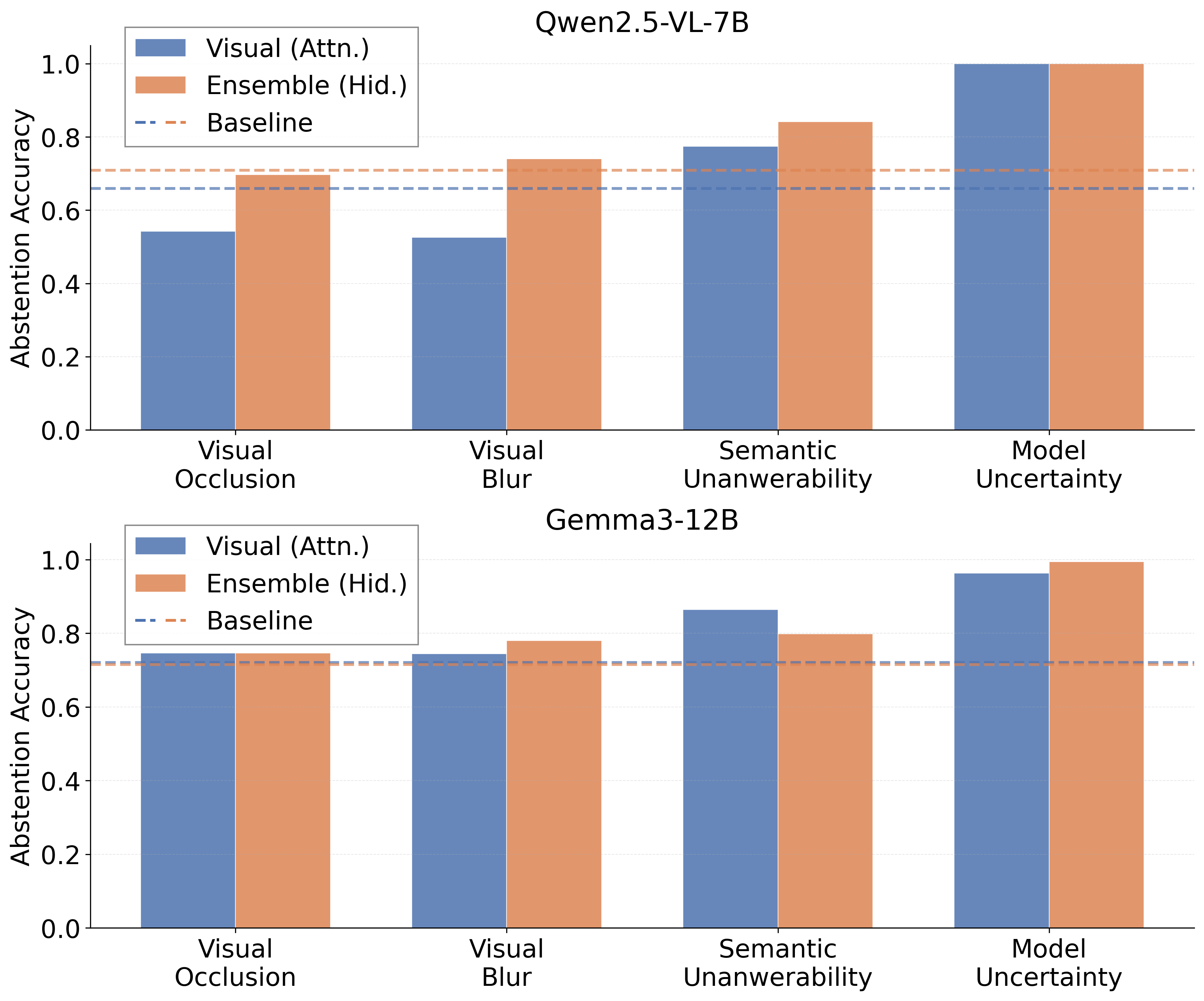}
    \caption{Abstention accuracy evaluated by probes trained on \textsc{HierText}. Baseline represents probes' abstention accuracy on original \textsc{HierText} dataset. LRP demonstrates high generalization capability to various unconfident scenarios, indicating it learns general confidence representation.}
    \label{fig:uncertain}
\end{figure}

Results in Figure \ref{fig:uncertain} shows that visual-focused attention probes achieve higher abstention accuracy than \textsc{HierText} where the probes are trained on in 6 out of 8 (model, dataset) combinations, while ensemble hidden state probes exceed in 7 out of 8 combinations. This cross-uncertainty generalization indicates that probes capture fundamental confidence signals rather than dataset-specific patterns. Furthermore, these results confirm that uncertainty signals are indeed encoded in the model's internal representations and can be effectively extracted through probing.

\subsection{Generalization}
\begin{table}[!t]
\centering
\tabcolsep=0.1cm
\small
\begin{tabular}{llccccc}
\toprule[1.5pt]
\makecell[l]{\textbf{Original} \\ \textbf{Dataset}} & \textbf{Method} & \textsc{OCR} & \textsc{Seed} & \textsc{Hier} & \textsc{Ego} & Avg\\
\midrule[0.75pt]\multicolumn{7}{c}{\textit{\ \ \textbf{\textsc{Qwen2.5-VL-7B}}} }\\
\midrule[0.75pt]
\multirow{4}{*}{\textsc{OCR}} & Self Consist. &0.677 &0.623 &0.625 &0.644 & 0.631\\
& VLM Judge &0.768 &\textbf{0.656} &0.632 &0.628 & 0.639\\
& Visual (Attn.) &\cellcolor{blue!10}0.768 &\cellcolor{blue!10}0.647 &\cellcolor{blue!10}0.613 &\cellcolor{blue!10}0.677 &\cellcolor{blue!10}0.646\\
& Ensemble (Hid.) &\cellcolor{blue!10}\textbf{0.789} &\cellcolor{blue!10}0.654 &\cellcolor{blue!10}\textbf{0.669} &\cellcolor{blue!10}\textbf{0.680} &\cellcolor{blue!10}\textbf{0.668}\\
\midrule[0.75pt]
\multirow{4}{*}{\textsc{Seed}} & Self Consist. &\textbf{0.709} &0.636 &\textbf{0.669} &0.662 & \textbf{0.680}\\
& VLM Judge &0.601 &0.737 &0.551 &0.632 & 0.595\\
& Visual (Attn.) &\cellcolor{blue!10}0.553 &\cellcolor{blue!10}0.735 &\cellcolor{blue!10}0.604 &\cellcolor{blue!10}\textbf{0.645} &\cellcolor{blue!10}0.601\\
& Ensemble (Hid.) &\cellcolor{blue!10}0.501 &\cellcolor{blue!10}\textbf{0.763} &\cellcolor{blue!10}0.563 &\cellcolor{blue!10}0.547 &\cellcolor{blue!10}0.537\\
\midrule[0.75pt]
\multirow{4}{*}{\textsc{Hier}} & Self Consist. &0.677 &0.623 &0.625 &0.644 & \textbf{0.648}\\
& VLM Judge &0.672 &\textbf{0.645} &\textbf{0.749} &0.628 & \textbf{0.648}\\
& Visual (Attn.) &\cellcolor{blue!10}0.676 &\cellcolor{blue!10}0.579 &\cellcolor{blue!10}0.659 &\cellcolor{blue!10}\textbf{0.668} &\cellcolor{blue!10}0.641\\
& Ensemble (Hid.) &\cellcolor{blue!10}\textbf{0.680} &\cellcolor{blue!10}0.568 &\cellcolor{blue!10}0.709 &\cellcolor{blue!10}0.645 &\cellcolor{blue!10}0.631\\
\midrule[0.75pt]
\multirow{4}{*}{\textsc{Ego}} & Self Consist. &\textbf{0.677} &0.623 &\textbf{0.625} &0.644 & \textbf{0.642}\\
& VLM Judge &0.619 &\textbf{0.715} &0.560 &0.634 & 0.632\\
& Visual (Attn.) &\cellcolor{blue!10}0.623 &\cellcolor{blue!10}0.689 &\cellcolor{blue!10}0.604 &\cellcolor{blue!10}0.722 &\cellcolor{blue!10}0.639\\
& Ensemble (Hid.) &\cellcolor{blue!10}0.587 &\cellcolor{blue!10}0.658 &\cellcolor{blue!10}\textbf{0.625} &\cellcolor{blue!10}\textbf{0.738} &\cellcolor{blue!10}0.623\\
\midrule[0.75pt]\multicolumn{7}{c}{\textit{\ \ \textbf{\textsc{Gemma3-12B}}} }\\
\midrule[0.75pt]
\multirow{4}{*}{\textsc{OCR}} & Self Consist. &0.743 &\textbf{0.726} &\textbf{0.771} &0.587 &0.694 \\
& VLM Judge &0.760 &0.607 &0.690 &0.615 &0.637 \\
& Visual (Attn.) &\cellcolor{blue!10}0.781 &\cellcolor{blue!10}0.684 &\cellcolor{blue!10}0.721 &\cellcolor{blue!10}- &\cellcolor{blue!10}\textbf{0.703} \\
& Ensemble (Hid.) &\cellcolor{blue!10}\textbf{0.797} &\cellcolor{blue!10}0.697 &\cellcolor{blue!10}0.697 &\cellcolor{blue!10}\textbf{0.622} &\cellcolor{blue!10}0.672\\
\midrule[0.75pt]
\multirow{4}{*}{\textsc{Seed}} & Self Consist. &0.720 &0.735 &\textbf{0.724} &0.610 &\textbf{0.685} \\
& VLM Judge &\textbf{0.733} &0.667 &0.666 &0.619 &0.673 \\
& Visual (Attn.) &\cellcolor{blue!10}0.536 &\cellcolor{blue!10}0.708 &\cellcolor{blue!10}0.613 &\cellcolor{blue!10}- &\cellcolor{blue!10}0.575 \\
& Ensemble (Hid.) &\cellcolor{blue!10}0.695 &\cellcolor{blue!10}\textbf{0.748} &\cellcolor{blue!10}0.508 &\cellcolor{blue!10}\textbf{0.629} &\cellcolor{blue!10}0.611 \\
\midrule[0.75pt]
\multirow{4}{*}{\textsc{Hier}} & Self Consist. &\textbf{0.743} &\textbf{0.726} &\textbf{0.771} &0.587 &\textbf{0.685} \\
& VLM Judge &0.657 &0.623 &0.632 &\textbf{0.613} &0.631 \\
& Visual (Attn.) &\cellcolor{blue!10}0.692 &\cellcolor{blue!10}0.601 &\cellcolor{blue!10}0.721 &\cellcolor{blue!10}- &\cellcolor{blue!10}0.647 \\
& Ensemble (Hid.) &\cellcolor{blue!10}0.633 &\cellcolor{blue!10}0.583 &\cellcolor{blue!10}0.715 &\cellcolor{blue!10}0.542 &\cellcolor{blue!10}0.586 \\
\midrule[0.75pt]
\multirow{4}{*}{\textsc{Ego}} & Self Consist. &\textbf{0.741} &\textbf{0.741} &\textbf{0.737} &\textbf{0.608} &\textbf{0.740} \\
& VLM Judge &0.543 &0.607 &0.495 &0.607 &0.549 \\
& Visual (Attn.) &\cellcolor{blue!10}- &\cellcolor{blue!10}- &\cellcolor{blue!10}- &\cellcolor{blue!10}- &\cellcolor{blue!10}- \\
& Ensemble (Hid.) &\cellcolor{blue!10}0.590 &\cellcolor{blue!10}0.590 &\cellcolor{blue!10}0.551 &\cellcolor{blue!10}0.647 &\cellcolor{blue!10}0.577 \\
\bottomrule[1.5pt]
\end{tabular}
\caption{Cross-dataset abstention accuracy. Each row shows results when training on the original dataset (left) and testing on different datasets (columns). ``OCR'', ``Seed'', ``Hier'', ``Ego'' are short for \textsc{OCRBench v2}, \textsc{SeedBench Plus 2}, \textsc{HierText} and \textsc{EgoTextVQA} respectively. Best results are in \textbf{bold}, our methods are in \colorbox{blue!10}{blue background}. ``Avg'' is the average abstention accuracy on unseen datasets (self-excluded). Self-consistency, a training-free abstention mechanism shows the best overall generalization capability, while visual-focused attention pattern probe generalizes best among training-based methods.}
\label{tab:generalization}
\end{table}
Prior work has noted that training-based abstention methods often suffer from limited generalization to unseen distributions \citep{feng2024don}. To evaluate this concern, we conduct cross-dataset evaluation by training probes on one dataset and testing on another, comparing our approach against the best-performing training-based baseline (VLM Judge) and threshold-based baseline (Self-Consistency) from Table \ref{tab:main}. 

Results in Table \ref{tab:generalization} show that self-consistency demonstrates better generalization across datasets. Among training-based methods, visual-focused attention probe achieves the best generalization with 0.636 average abstention accuracy, outperforming best training baseline VLM Judge. Attention probes generalize better than hidden state probes due to their simpler, more universal representations.

However, training data diversity significantly impacts generalization. When trained on \textsc{OCRBench v2}, the most comprehensive dataset, visual-focused attention probe achieves 0.675 average cross-dataset abstention accuracy, surpassing self-consistency by 1.8\% relative improvement. This demonstrates that with sufficient training data covering diverse uncertainty patterns, learned probes can achieve strong generalization capabilities.

\subsection{Optimal Layer}
\begin{figure}[!t]
    \centering
    \includegraphics[width=1\linewidth]{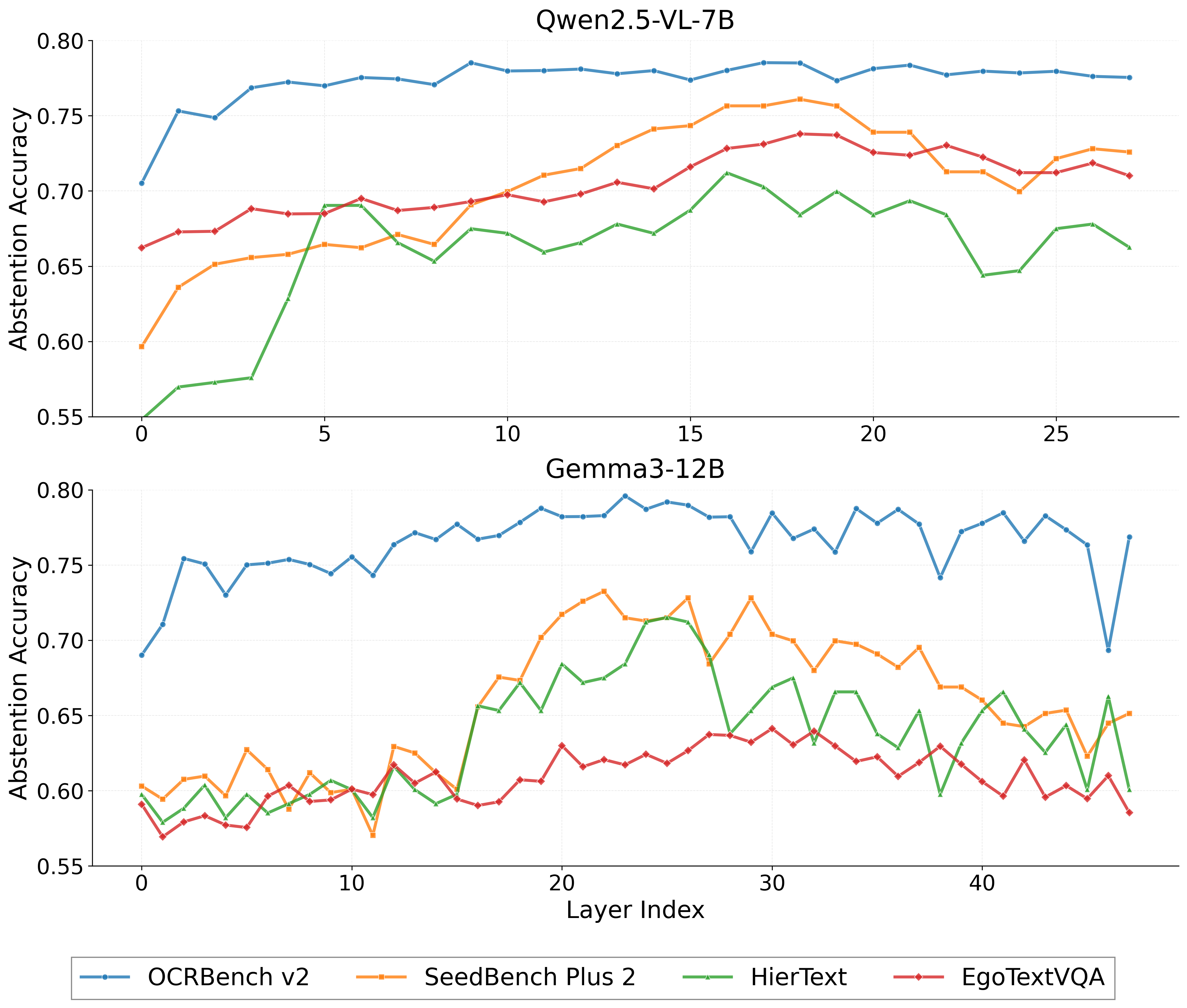}
    \caption{The abstention accuracy of probes trained on each layer's hidden states across four datasets. For \textsc{SeedBench}, \textsc{HierText} and \textsc{EgoTextVQA}, the abstention accuracy first increases and then decreases, indicating that confidence signals are richer in the intermediate instead of final layers.}
    \label{fig:layer}
\end{figure}

To understand why latent representations encode richer confidence information, we analyze how confidence signals evolve across transformer layers. We plot the abstention accuracy of probes trained on each layer's hidden states across four datasets in Figure \ref{fig:layer}. 

We observe two key patterns. For \textsc{SeedBench Plus 2}, \textsc{HierText}, and \textsc{EgoTextVQA}, abstention accuracy exhibits a clear rise-then-fall trend, peaking around layers 15-20 for \textsc{Qwen2.5-VL-7B} and around layers 20-30 for \textsc{Gemma3-12B}. For \textsc{OCRBench v2}, accuracy remains consistently high across most layers after initial rise. Notably, \textsc{Gemma3-12B} shows noisier patterns with confidence peaks appearing in shallower layers compared to \textsc{Qwen2.5-VL-7B}, which may reflect insufficient training or over-aggressive alignment given its lower performance.

The rise-then-fall pattern confirms that confidence signals are strongest in intermediate layers, validating prior findings that alignment processes degrade calibration near output layers \citep{wu2023language, kossen2024semantic}. However, \textsc{OCRBench v2}'s consistent performance demonstrates that with sufficient training data diversity, probes can extract effective confidence signals even from later layers where signals are weaker. This reduces the pressure of layer selection when adequate training data is available.

\section{Conclusion}

In this work, we address a critical gap in VLM deployment: the ability to abstain when uncertain in Scene Text Visual Question Answering tasks. We propose Latent Representation Probing, a lightweight approach that trains probes on latent representations across transformer layers to detect low confidence. We provide the first comprehensive evaluation of state-of-the-art abstention mechanisms on VLMs. Experiments demonstrate that LRP substantially outperforms baselines by 7.6\% in abstention accuracy, while requiring only a single forward pass. Our analysis reveals three key insights: (1) LRP learn generalizable uncertainty representations that transfer across distinct unconfident scenarios, achieving 0.825 average abstention accuracy; (2) LRP exhibits superior cross-dataset generalization, outperforming training-based baselines by 5.8\% in abstention accuracy; (3) optimal confidence signals emerge in intermediate layers rather than final layers. Our paper reveals a broader principle: model confidence is not merely an output property but emerges progressively through internal computation, offering a principled path toward deployment-ready AI systems.

{
    \small
    \bibliographystyle{ieeenat_fullname}
    \bibliography{main}
}
\clearpage
\setcounter{page}{1}
\maketitlesupplementary

\section{Experiment Details}
\label{appendix:exp_detail}

\paragraph{Model Details} We employ 2 LLMs in the experiments, the detailed information of LLMs we used in this paper is as follows: 1) \textsc{Qwen2.5-VL-7B}, through the \href{https://huggingface.co/Qwen/Qwen2.5-VL-7B-Instruct}{\textsc{Qwen/Qwen2.5-VL-7B-Instruct}} checkpoint on Huggingface \citep{wolf2020transformers}; 2) \textsc{Gemm3-12B}, through the \href{https://huggingface.co/google/gemma-3-12b-it}{\textsc{google/gemma-3-12b-it}} checkpoint on Huggingface.

\paragraph{Dataset Details} We employ 4 datasets in the experiments. The detailed information is:

\begin{itemize}[leftmargin=*]
    \item \textbf{OCRBench v2 (OCR)} \citep{fu2024ocrbench}. We collect the dataset through the \href{https://huggingface.co/lmms-lab/OCRBench-v2}{\textsc{lmms-lab/OCRBench-v2}} checkpoint on Huggingface. We randomly split the official test split (10,000 questions) to train, validation and test subsets with 6:2:2 ratio.
    \item \textbf{SeedBench Plus 2 (Seed)} \citep{li2024seed}. We collect the dataset through the \href{https://huggingface.co/AILab-CVC/SEED-Bench-2-plus}{\textsc{AILab-CVC/SEED-Bench-2-plus}} checkpoint on Huggingface. We randomly split the official test split (2,277 questions) to train, validation and test subsets with 6:2:2 ratio.
    \item \textbf{HierText (Hier)} \citep{long2022towards}. We collect the dataset through the \href{https://github.com/google-research-datasets/hiertext}{\textsc{google-research-datasets/hiertext}} repository on Github. We randomly split the official test split (1,612 questions) to train, validation and test subsets with 6:2:2 ratio. For each image, \textsc{Gemini-2.5-Pro} will first randomly choose from paragraph, line or word perspective to synthesize the question. Then it will generate a QA pair with reference to the ground-truth OCR annotation results. Last, it will replace a few words in the original to make up an unanswerable question which ask about a non-existing region or text. The full prompt we use for generating the questions are in Table \ref{tab:prompt}:
    \item \textbf{EgoTextVQA (Ego)} \citep{zhou2025egotextvqa}. We collect the dataset through the \href{https://github.com/zhousheng97/EgoTextVQA}{\textsc{zhousheng97/EgoTextVQA}} repository on Github. We randomly split the official outdoor split (4,848 questions) to train, validation and test subsets with 6:2:2 ratio. The indoor split is not inlcuded due to computtaion limitation.
\end{itemize}
\begin{table*}[!htbp]
\centering
\begin{tabular}{p{0.95\linewidth}}
\toprule[1.5pt]
Generate TWO questions based on this image and ground-truth OCR results:
\begin{enumerate}
    \item \textbf{ANSWERABLE QUESTION}: A question that CAN be answered using the visible text in the image.
    \begin{itemize}
        \item The question must reference text location using spatial descriptions. Be diversified and creative about the spatial descriptions (including but not limited to, ``the paragraph at the top-left'', ``the second line in the middle section'', ``the word next to the logo'', etc.).
        \item The answer must be exactly one OCR result (paragraph, line, word) from the ground-truth OCR results with the exact corresponding vertices.
        \item Make sure the answer is unique and unambiguous from the question with enough constraints.
    \end{itemize}
    
    \item \textbf{UNANSWERABLE QUESTION}: A question that CANNOT be answered from the visible text alone.
    \begin{itemize}
        \item Created by simply replacing a few words from the answerable question.
        \item The question is unanswerable because it clearly references an invalid text location (region with no text or doesn't exist) in the image.
    \end{itemize}
\end{enumerate}

\textbf{Output Format}:

\begin{lstlisting}
{
  "answerable_question": {
    "question": "Your question here",
    "answer": "Ground-truth answer here",
    "vertices": [[x1, y1], [x2, y2], ..., [xn, yn]]
  },
  "unanswerable_question": {
    "question": "Your question here"
  }
}
\end{lstlisting}
\\
\bottomrule[1.5pt]
\end{tabular}
\vspace{5pt}
\caption{System prompt for synthesizing \textsc{HierText} questions.} \label{tab:prompt}
\end{table*}

\paragraph{Inference and Training Details}
For all experiments, we use greedy decoding (temperature = 0.0 and top\_p == 1.0) and set max\_new\_tokens=128. For evaluation, we first utilize \textsc{Gemini 2.5 Pro} to parse the responses into containing only the answer part to get rid of the possible redundant introduction like ``Sure! The answer is'' and then leverage the official evaluation suite get the response accuracy as labels. We use the standard SFTTrainer in Huggingface for VLM fine-tuning. We use grid search on learning rate, weight decay and learning rate scheduler and early stop mechanism to find the best hyperparameter for probe-based methods.
\citep{yao2025mmmg}

\end{document}